\documentclass[journal]{IEEEtran}
\ifCLASSINFOpdf
\else
\fi
\usepackage{bbold}
\usepackage{ulem}
\usepackage{tikz}
\usepackage{microtype}
\usepackage{graphicx}
\usepackage{hyperref}
\usepackage{amsmath}
\usepackage{amssymb}
\usepackage{mathtools}
\usepackage{amsthm}
\usepackage{subcaption}
\usepackage{diagbox}

\usepackage{amsfonts}
\usepackage{fancyhdr}
\usepackage{amssymb}
\usepackage{xcolor} 
\definecolor{Prune}{RGB}{99,0,60}
\usepackage{mdframed}
\usepackage{multirow} 
\usepackage{multicol} 
\usepackage{scrextend} 
\usepackage{tikz}
\usepackage{graphicx}
\usepackage[absolute]{textpos} 
\usepackage{colortbl}
\usepackage{array}
\usepackage{geometry}
\usepackage{hyperref}
\usepackage{bbold}
\usepackage{ulem}
\usepackage{amsthm}
\usepackage{stmaryrd}
\usepackage{algpseudocode} 
\usepackage{stmaryrd}
\usepackage{caption}
\usepackage{subcaption}
\usepackage{tablefootnote}
\usepackage{algorithm} 
\usepackage{upgreek}

\newcommand{\addSylv}[1]{\textcolor{black}{#1}}
\newcommand{\addSido}[1]{\textcolor{black}{#1}}

\begin{document}
%
\title{Self-Supervised Learning for Real-World Object Detection: a Survey}
%
%
%

\author{Alina Ciocarlan,
        Sidonie Lefebvre,
        \addSylv{Sylvie Le Hégarat-Mascle}
        and~Arnaud Woiselle}

\maketitle

\begin{abstract}
Self-Supervised Learning (SSL) has emerged as a promising approach in computer vision, enabling networks to learn meaningful representations from large unlabeled datasets. SSL methods fall into two main categories: instance discrimination and Masked Image Modeling (MIM). While instance discrimination is fundamental to SSL, it was originally designed for classification and may be less effective for object detection, particularly for small objects. In this survey, we focus on SSL methods specifically tailored for real-world object detection, with an emphasis on detecting small objects in complex environments. Unlike previous surveys, we offer a detailed comparison of SSL strategies, including object-level instance discrimination and MIM methods, and assess their effectiveness for small object detection using both CNN and ViT-based architectures.
Specifically, our benchmark is performed on the widely-used COCO dataset, as well as on a specialized real-world dataset focused on vehicle detection in infrared remote sensing imagery. We also assess the impact of pre-training on custom domain-specific datasets, highlighting how certain SSL strategies are better suited for handling uncurated data.

Our findings highlight that instance discrimination methods perform well with CNN-based encoders, while MIM methods are better suited for ViT-based architectures and custom dataset pre-training. This survey provides a practical guide for selecting optimal SSL strategies, taking into account factors such as backbone architecture, object size, and custom pre-training requirements. Ultimately, we show that choosing an appropriate SSL pre-training strategy, along with a suitable encoder, significantly enhances performance in real-world object detection, particularly for small object detection in frugal settings.

\end{abstract}


%
\IEEEpeerreviewmaketitle

\section{Introduction}
\label{sec:intro}

\begin{figure*}[]
         \centering
         \includegraphics[width=15cm]{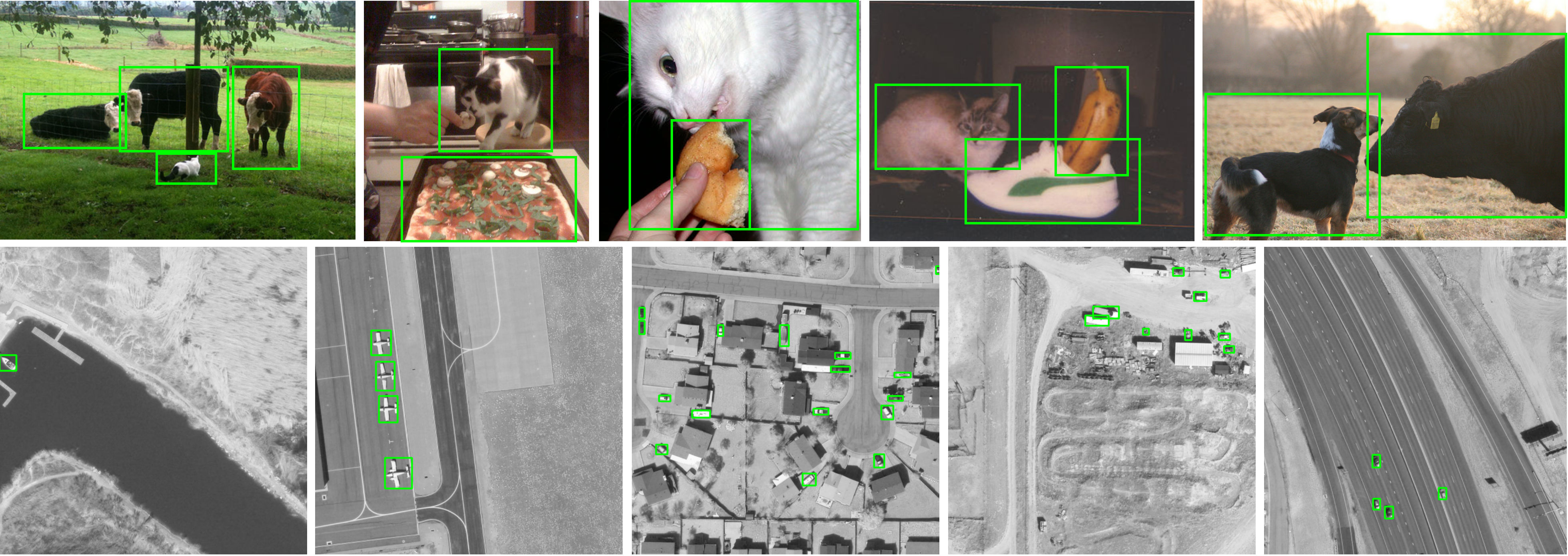}
         \caption{Example of images dealing with object detection. The first row  \addSylv{shows some} images taken from the COCO dataset~\cite{lin2014microsoft}, and the second row provides some infrared images taken from the VEDAI dataset~\cite{razakarivony2016vehicle}. Objects are framed in green.}
         \label{fig:ex_det_obj}
\end{figure*}

Self-supervised learning (SSL) is an exciting and active research area in computer vision. It consists in an unsupervised training of deep learning networks (often only the encoder) \addSylv{using} a well-designed pretext task. The aim of this pre-training task is 
to help the network learning features or invariances that are relevant for the downstream task. \addSylv{In the literature,} SSL methods have been shown to improve SOTA performance for many use cases. More specifically, SSL allows the network to learn general features from large unlabelled datasets which, when transferred to a final task, will improve performance despite difficult fine-tuning conditions (e.g., little annotated data or few computational resources). 

Fundamental SSL methods deal with instance discrimination, which aims at modeling the decision borders between sub-sets of data represented in the latent space. These methods consider images as instances, and perform inter-image discrimination. Concretely, the optimization \addSylv{aims to minimize}, in the latent space, the distance between features of instances that share similar semantic properties (e.g., augmented views from the same anchor images). Emblematic methods include MoCov2~\cite{he2020momentum}, BYOL~\cite{grill2020bootstrap} and DINO~\cite{caron2021emerging}. We refer the reader to the following surveys~\cite{jaiswal_survey_2021, ozbulak2023know, gui2024survey} for more details about instance discrimination methods.


However, instance discrimination methods \addSylv{were primarily} designed for classification tasks, and most of them are benchmarked on classification datasets only. 
Although some methods~\cite{he2020momentum,he2022masked} provide promising results on famous object detection datasets like COCO~\cite{lin2014microsoft} or ADE20K~\cite{zhou2019semantic}, they were not \addSylv{specifically} designed for object detection and thus \addSylv{may} appear sub-optimal for this task, and even worse for small object detection. This is especially \addSylv{true} for instance discrimination methods that \addSylv{mostly involve} inter-image comparisons, assuming that the images are semantically consistent. To overcome this problem, some object-level instance discrimination methods have been proposed in the literature~\cite{xiao2021region, wang2021dense,xie2021propagate,henaff2022object,wang2022exploring,ziegler2022self}. They either rely on local crops to create positive pairs, or on dense instance discrimination loss. \addSylv{Then, a}nother recently introduced SSL paradigm that deals with local feature analysis is Masked Image Modeling (MIM). Unlike instance discrimination, MIM methods naturally deal with modeling local relationships: neighboring pixels are all the more important to reconstruct masked patches. 

Although object-level instance discrimination and MIM methods have been shown to \addSylv{be efficient for} local or dense prediction task\addSylv{s}, it remains unclear which paradigm is better suited for object detection. Few studies have attempted to compare instance discrimination and MIM paradigms~\cite{li2021benchmarking, park2022self, xie2023revealing, gao2024observation}. These studies all agree that MIM methods lead to better performance than instance discrimination methods when fine-tun\addSylv{ed} on data-sufficient object detection dataset. Specifically, \addSylv{the authors of}~\cite{xie2023revealing} observe that MIM shows a local inductive bias at all layers while MoCov3 (\addSylv{that is an} instance discrimination \addSylv{method}) tends to focus on local details in lower layers and on global details in higher layers. They also show that MIM pre-training brings sufficient diversity \addSylv{to} the attention heads, unlike instance discrimination pre-training \addSylv{strategies} \addSylv{whose} capacity may thus be limited. The authors conclude that coupling MIM methods with Vision Transformer (ViT) encoders should lead to state-of-the-art (SOTA) performance. However, \cite{gao2024observation} extends these studies by evaluating these methods in data-limited contexts, \addSylv{and} conclud\addSylv{es} that while MIM methods often outperform contrastive learning methods on large downstream datasets, they struggle with smaller datasets.

These surveys have two key limitations: 1)~they do not consider object-level instance discrimination methods, and 2)~they rely exclusively on ViT backbones. As a result, the conclusions have not been validated on CNN-based backbones such as ResNets. CNN-based encoders remain widely used in many real-world applications and have some advantages, such as faster inference times, and a hierarchical architecture that benefits object detection. \addSylv{The authors of}~\cite{huang2022survey} evaluate some local instance discrimination methods using a ResNet-50 backbone, but they only considered a few-shot setting and did not compare with MIM methods. Moreover, the results were directly taken from the original papers, which, as the authors noted, could lead to unfair comparisons due to differences in implementation. 

We aim to address these gaps by providing a comprehensive survey of SSL methods tailored for object detection, with a focus on challenging cases such as small objects or frugal contexts. Specifically, we extensively cover local instance discrimination and MIM methods. We then benchmark a selection of methods, ensuring \addSylv{the representativeness} across all SSL categories and network architectures. We compare global, local instance discrimination, and MIM methods using two network sizes, considering both ResNet-50 and ViT backbones. Our first benchmark uses the widely recognized COCO dataset, with a particular focus on the performance on small objects. We then move to a real-world application involving small object detection, namely vehicle detection from remote sensing data, using the VEDAI dataset~\cite{razakarivony2016vehicle}. Some examples of images taken from both datasets are provided in Figure~\ref{fig:ex_det_obj}. An important \addSylv{limitation} of previous studies is their primary focus on the COCO dataset, which is not representative of many real-world object detection scenarios. In practical applications, objects may be very different (e.g., very small) and hidden within complex backgrounds. Additionally, different sensors may be used, such as hyperspectral sensors, making pre-trained weights on RGB images less applicable. In such cases, it is necessary to train SSL methods on a dataset that shares similar spectral characteristics with the target dataset. The quality of SSL pre-training (i.e., pre-training that leads to high fine-tuning performance) and the choice of SSL method will then heavily depend on the characteristics of the pre-training dataset (e.g., temporal redundancy, image diversity, dataset size, etc.). We therefore propose \addSylv{an experiment where we} pre-train on a large-scale, non-curated IR dataset and \addSylv{we} evaluat\addSylv{e} the benefits of custom pre-training on the IR version of the VEDAI dataset compared to using weights pre-trained on RGB images.

Our contributions can be summarized as follows: 
\begin{itemize} 
\item We provide an exhaustive survey of SSL methods tailored for object detection, with a specific focus on local instance discrimination and MIM methods. 
\item We evaluate representative SSL methods from each category on two benchmarks. First, we consider the widely used COCO dataset, emphasizing metrics related to small objects. Then, we evaluate these SSL strategies in a real-world application, specifically vehicle detection from remote sensing data. This allows us to draw conclusions on the optimal SSL strategy depending on various parameters (ResNet or ViT backbone, object size, fine-tuning dataset size, etc.). 
\item We offer insights on which SSL strategy to use when pre-training on an in-domain dataset is required. 
\end{itemize}

\section{Towards local-level self-supervised learning}
In \addSylv{this section}, we present some SSL strategies that are better suited to dense or local prediction tasks (e.g., segmentation and object detection, respectively) as they aim to learn local features. They can be grouped within two categories, namely object-level instance discrimination methods and masked image modeling. Table~\ref{tab:taxo_ssl_general} summarizes the different categories and the associated methods \addSylv{that} will be discussed. 

\bgroup
\def\arraystretch{1.15}
\begin{table*}[ht] 
\small
\centering
  \begin{tabular}{p{2.8cm}p{1.8cm}lp{5.5cm}} 
  \hline
  
\hline


    \multirow{6}{2.8cm}{\textbf{Object-level instance discrimination}} &  \multicolumn{2}{l}{\textit{Region-level augmentations}} & SCRL, \textbf{ReSim}, MaskCo, SoCo, CAST, ContrastiveCrop, InsLoc, CP², ORL, \textbf{Leopart}, InsCon \\
    
    \arrayrulecolor{black}\cline{2-4}
    & \multirow{3}{2cm}{\textit{Dense loss}} & Raw pixels & VaDeR, PixContrast, PixPro, DUPR, InsCon, \textbf{Leopart}, LC-Loss, CLOVE \\
    \arrayrulecolor{lightgray}\cline{3-4}
    & & Feature matching & DenseCL, Self-EMD, VicRegL \\
    \arrayrulecolor{lightgray}\cline{3-4}
    & & Semantic alignment & DetCon, Odin, SetSim\\
    \arrayrulecolor{black}
\hline

\hline
    \multirow{4}{2.8cm}{\textbf{Masked Image Modeling}} & \multicolumn{2}{l}{\textit{Specific masking strategy}} &  CIM, MST, AttMask, AMT, MILAN, SemMAE, DPPMask, MixMAE \\
    \arrayrulecolor{black}\cline{2-4}
    & \multirow{3}{2cm}{\textit{Target objective}} & Geometric alignment & \textbf{MAE}, SimMIM,  ConvMAE, \textbf{SparK} \\
    \arrayrulecolor{lightgray}\cline{3-4}
    & & Image descriptors & PixMIM, Ge²-AE, A²MIM, MaskFeat, SSM \\
    \arrayrulecolor{lightgray}\cline{3-4}
    & & Deep features & BEiT, MaskDistill, MILAN, MaskAlign, SplitMask, iBOT, I-JEPA, dBOT\\
   \arrayrulecolor{black}\hline
   
\hline

  \end{tabular}
  \caption{Taxonomy of local-level SSL methods for image representation learning. The methods we will consider in our experiments are \addSylv{shown} in bold.}
  \label{tab:taxo_ssl_general}
\end{table*}
\egroup
\def\arraystretch{1.15}

\subsection{Object-level instance discrimination methods}
Some authors propose\addSylv{d} variants of instance discrimination methods that are well suited \addSylv{to} object detection tasks. 
Instance discrimination methods aim at minimizing the distance \addSylv{in the latent space} between features of instances that share similar semantic properties. \addSylv{The f}undamental methods presented in the Introduction perform \addSylv{only} inter-image comparison\addSylv{s}: they generally consider the entire images as their instances, assuming that these are semantically consistent. This is indeed the case when the methods are trained on object-centric datasets such as ImageNet. However, this hypothesis does not necessarily hold when dealing with dense prediction tasks such as object detection or segmentation. To overcome this issue, two approaches have been \addSylv{investigated}: designing \addSylv{data-augmentations at the} object or region-level, or applying instance discrimination loss at a local-level (e.g., per pixel).

 

  
\subsubsection{Region-level augmentations}



\addSylv{The approach} consists in applying instance discrimination loss \addSylv{to} local patches in order to perform intra-image instance discrimination. Several strategies have been proposed to ensure semantic consistency between images \addSylv{that form} a positive pair. 
\begin{figure*}[h]
         \centering
         \includegraphics[width=14.5cm]{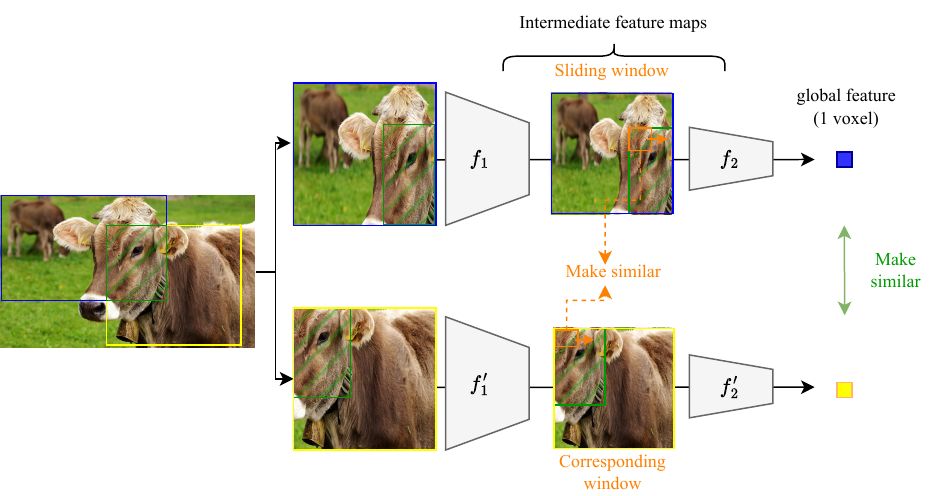}
         \caption{Example of object-level instance discrimination pipeline. Here, we represented the ReSim framework, which consists in maximizing the similarity between a sliding window in the first branch and its equivalent in the second branch, within an overlapping area.}
         \label{fig:resim}
     \end{figure*}
\addSylv{Spatially Consistent Representation Learning (SCRL)}~\cite{roh2021spatially} first propose\addSylv{d} to randomly select boxes within the intersecting area of the two positive samples and to minimize the similarity between the features predicted by the pooled boxes. Concurrently, \cite{xiao2021region} proposed a similar approach called ReSim. As shown in Figure~\ref{fig:resim}, a sliding window extracts, in each branch, local features within the overlapping area between the two augmented views of the anchor sample (dashed green area). This creates local positive pairs that represent exactly the same spatial region in the original image (we say that the patches are geometrically aligned).
Unlike SCRL, the loss is applied at three different scales in the network, which benefits the detection of objects of different size\addSylv{s}. ReSim also performs inter-image instance discrimination between the two global features (representing the entire positive sample) extracted by the network in order to \addSylv{maintain} good performance \addSylv{i}n classification tasks. MaskCo~\cite{zhao2021self} further introduces the Contrastive Mask Prediction task. It consists in masking one of the local patches (query patch, taken from the first branch), and predict\addSylv{ing} which augmented view (key views from the second branch) suits the best to fill the masked query patch. Negative key views are introduced by randomly sampling patches from the rest of the dataset, and the contrastive loss is applied to perform the Contrastive Mask Prediction task.

Nevertheless, SCRL, ReSim and MaskCo assume that \addSylv{all} overlapping area\addSylv{s are} 
semantically consistent, which may not be the case on dense visual scenes (e.g., if the size of the overlapping area
is too large). To avoid this issue, SoCo~\cite{wei2021aligning} \addSylv{relies on} the selective search algorithm used in Faster R-CNN to extract semantically consistent sub-regions of an image. 
Furthermore, CAST~\cite{selvaraju2021casting} introduces saliency random crop\addSylv{ping}. Saliency maps are learned with Grad-CAM supervision, and their goal is to identify foreground objects (and thus semantically consistent regions) within an anchor image. ContrastiveCrop~\cite{peng2022crafting} goes further and proposes not only a semantic-aware crop\addSido{ping} based on the heatmap analysis during the contrastive training, but also a centre-suppressed sampling (i.e., by limiting center crops) that increases the variance in the crops. Indeed, one issue with random crops is that \addSylv{they may} introduce \addSylv{too} easy positive pairs. 
\addSylv{Then}, InsLoc~\cite{yang2021instance} and $\mbox{CP}^2$~\cite{wang2022cp} introduce background invariance in\addSylv{to} their crops by copy\addSylv{ing}-pasting foreground images (e.g., crops from ImageNet dataset) on different background images. In their loss, they ensure that \addSylv{the} features extracted for the pasted foreground object are similar, regardless of the background. 

However\addSylv{,} all the methods presented so far rely on intra-image positive pairs, which limits the diversity of information contained in positive pairs. \addSylv{Object-level Representation Learning (}ORL\addSylv{)}~\cite{xie2021unsupervised} addresses this issue by relying on a three-stage pipeline. First, an instance discrimination method (e.g., BYOL) \addSylv{is trained} on \addSylv{an} object-centric dataset (e.g., ImageNet) to learn to extract global features. Second, the pre-trained encoder is used to \addSylv{generate} local positive pairs. For this purpose, global features are extracted on the target dataset using the pre-trained backbone, and similar images are clustered together using a \addSylv{K-Nearest Neighbors (}KNN\addSylv{) algorithm}. A selective search algorithm is then used to extract local regions within the similar images, and positive pairs of local patches are matched using the encoder pre-trained in the first step jointly with a KNN clustering. Third, another instance discrimination method \addSylv{is trained} using the newly \addSylv{generated} local positive pairs. Another \addSylv{alternative} is
to combine an instance discrimination method based on clustering, such as SwAV, and local augmentations. Leopart~\cite{ziegler2022self} builds upon this solution. More specifically, it consists in providing two crops of a foreground object (identified by leveraging ViT attention maps) to an instance discrimination network (e.g., DINO), and then producing patch-level cluster assignments, which are forced to be similar following the online optimization objective of SwAV~\cite{caron2020unsupervised}.
Finally, to improve multi-object detection, InsCon~\cite{yang2022inscon} ensures multi-instance consistency by taking as a query sample a multi-instance view containing four images, and as positive samples augmentations of each individual image contained in the query sample.

\subsubsection{Dense loss}

\begin{figure*}[h]
         \centering
         \includegraphics[width=14cm]{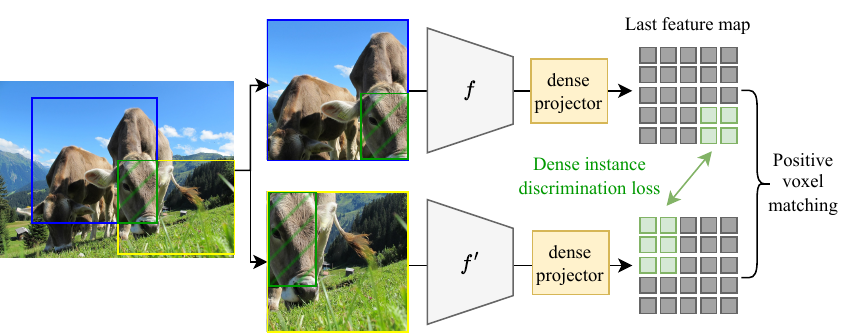}
         \caption{Dense instance discrimination loss.}
         \label{fig:dense_cl}
     \end{figure*}

The second idea for improving SSL for dense prediction tasks is to apply an instance discrimination loss at ``pixel'' level (i.e.\addSylv{,} each voxel of the last feature map), as illustrated on Figure~\ref{fig:dense_cl}. Such \addSylv{a} strategy boils down to dividing the image into a grid and taking all (or most \addSylv{of}) the patches in the grid into account when computing the instance discrimination loss. 
The key to this type of method lies in \addSylv{how} the positive voxel are matched, i.e. how the features of different views are aligned. In the literature, several alignment strategies \addSylv{have been} proposed:

\paragraph{Geometric alignment}  VaDeR\addSylv{~\cite{o2020unsupervised}}, PixContrast~\cite{xie2021propagate}, PixPro~\cite{xie2021propagate}, DUPR~\cite{ding2022deeply}, InsCon~\cite{yang2022inscon}, Leopart~\cite{ziegler2022self}, LC-Loss~\cite{islam2023self} and CLOVE~\cite{silva2023self} assume that the geometric transforms between the positive images are known (thanks to the know\addSylv{ledge of the} data-augmentation process), and use them \addSylv{to perform} spatial alignment. Leopart~\cite{ziegler2022self} \addSylv{additionally} relies on the attention maps provided by \addSylv{the} ViT encoder to focus \addSylv{only} on foreground objects in the loss. PixPro further ensures spatial smoothness by propagating the features from similar pixels. CLOVE proposes a similar approach but \addSylv{instead} relies on self-attention maps to propagate features.  
    
\paragraph{Learned feature matching} It is not always possible to access geometric correspondences, as for example in the case of temporal positive pairs. \addSylv{Therefore,} DenseCL~\cite{wang2021dense}, Self-EMD~\cite{liu2020self} and VicRegL~\cite{bardes2022vicregl} align feature voxels that have a minimal distance between their values. \addSylv{An} obvious issue with relying solely on feature alignment is that it assumes that the feature extraction is semantically meaningful, which is not the case at the beginning of the training. On the one hand, DenseCL proposes a warm-up before applying this strategy, \addSylv{al}though they show that random matching (i.e., no\addSylv{t} semantically consistent matching) also leads to good performance. On the other hand, VicRegL combines learned feature matching with spatial matching.
    
\paragraph{Semantic alignment} To ensure semantic consistency between positive pairs of voxels, DetCon~\cite{henaff2021efficient} estimates pixel categories (pseudo-label\addSylv{s}) through unsupervised segmentation masking (using Felzenszwalb-Huttenlocher algorithm~\cite{felzenszwalb2004efficient}). The \addSylv{authors} show empirically that more accurate segmentation masks lead to better fine-tuning performance. In the same line, Odin~\cite{henaff2022object} trains an object \textit{discovery} network \addSylv{together} with an instance discrimination pipeline. More specifically, the object \textit{discovery} network relies on K-means clustering to cluster the features in the latent space, assuming that each cluster \addSylv{is} more likely to represent an object as  the training process \addSylv{progresses}. Concurrently, SetSim~\cite{wang2022exploring} uses attention maps to estimate both positive pixels location and similar sets of pixels, and then computes the similarity between the sets of pixels.

\subsection{Masked Image Modeling}
Conversely to instance discrimination methods whose goal is to estimate \addSylv{some} decision borders between image representations,  MIM consists in masking a relatively high proportion of an image and reconstructing \addSylv{it} (or \addSylv{its} features). This brings occlusion invariance to the encoder, as well as locality inductive bias~\cite{xie2023revealing}. The underlying hypothesis is that if a network is able to guess or even reconstruct severely corrupted information, then it ``understands'' the semantics in the image. Well-known SSL SOTA pipelines such as BEiT~\cite{bao2021beit}, Masked AutoEncoders (MAE)~\cite{he2022masked}, iBOT~\cite{zhou2021image} or I-JEPA~\cite{assran2023self} rely on this principle. They differ \addSylv{mainly} in the considered masking strategy \addSylv{and} the reconstruction objectives.

\subsubsection{Masking strategy}

In the literature, it has been shown that fine-tuning performance \addSylv{is highly dependent} on the masking strategy. We propose to group them by answering the following questions:

\begin{figure*}[h]
         \centering
         \includegraphics[width=13cm]{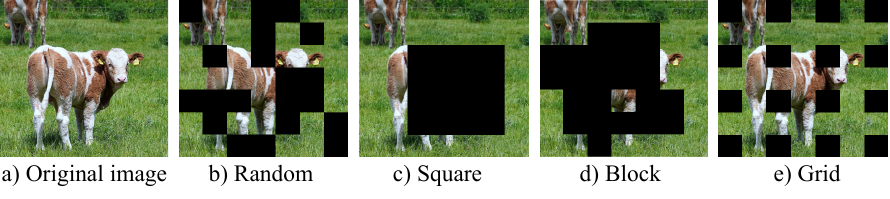}
         \caption{Common masking strategies for masked image modelling.}
         \label{fig:masking}
     \end{figure*}
    
\paragraph{What shape for the mask?}
    
Authors from MAE~\cite{he2022masked} and SimMIM~\cite{xie2022simmim} evaluate different mask sampling strategies that were previously proposed in the literature, including random, square~\cite{pathak2016context}, block-wise~\cite{bao2021beit} and grid masking strategies. The different masking strategies are represented on Figure~\ref{fig:masking}. Both works conclude that the simple random masking strategy is the most efficient, under the condition of considering a high masking ratio. Indeed, such \addSylv{a} strategy \addSylv{preserves} more hints about the object\addSylv{,} especially when considering an object-centric dataset, \addSylv{as opposed to the} square and blockwise strategies. 
Compared to \addSylv{the} regular grid masking strategy, random masking brings more difficult\addSylv{ies to} the network since \addSylv{the} object parts are \addSylv{unevenly} occluded. \addSylv{Therefore, a s}emantic understanding of non-occluded patches is necessary to reconstruct some \addSylv{heavily} occluded parts. A commonly chosen size for the masked patches is $32$ when considering pre-training on images of size $224\times 224$, which has shown to be efficient for many famous computer vision datasets (ImageNet~\cite{deng2009imagenet}, COCO~\cite{lin2014microsoft}, etc.). Note that such patch masking strategies (in terms of size and shape) may not be suitable for some real-world application and data, like in remote sensing or medical domains.\addSylv{The article~}\cite{li2023you} proposes masks with irregular shapes, which \addSylv{are} beneficial for anomaly detection in remote sensing images \addSylv{because} the authors simulate the spatial morphology of the anomalies. 
    
\paragraph{At which ratio?}

Masking a high ratio of patches is also important to make the pretext task difficult enough for the network\addSylv{, forcing it} to extract meaningful features. MAE and SimMIM \addSylv{have shown} that a ratio of $50\%$ is optimal for random masking. SimMIM further proposes a metric called Average Distance (\textit{AvgDist}) that evaluates the reconstruction difficulty of a given mask sampling strategy. It consists in computing the averaged Euclidean distance between masked pixels and the nearest visible ones. They conclude that masking strategies with an \textit{AvgDist} metric between $10$ and $20$ have more chance to perform well for fine-tuning. Note that this study has been performed on object-centric datasets (ImageNet, iNaturalist-2018~\cite{van2018inaturalist}), as well as on visual scenes (COCO and ADE20K~\cite{zhou2019semantic}).


\paragraph{Which values for the masked areas?}
    
    First transformer-based MIM methods propose to replace the masked patches by learnable embeddings~\cite{bao2021beit, xie2022simmim}. However, MAE showed that encoding masked patches leads to worse results: in addition to \addSylv{a significant impact on} the convergence time, it also brings a gap between pre-training and fine-tuning. Indeed, in the fine-tuning task, there are no such corrupted patches. Therefore, \addSylv{the} authors \addSylv{of} MAE paper propose to encode unmasked patches only, and design a specific decoder that takes as input the masked patches as learnable embeddings. 

Another strategy consists in replacing \addSylv{the} masked patches by plausible patches. CIM~\cite{fang2022corrupted} replaces the masking strategy by a more subtle corruption created using a generative network. Such masking strategy seems particularly appropriate for anomaly detection tasks, although the generation of subtle corruptions and their encrustation raise many questions.

\paragraph{Where?}

Some papers observe that masking patches at random locations can impair the performance of the network~\cite{liu2023pixmim}. Indeed, if the pre-training dataset contains small objects, they may be totally occluded. The objective of the network will therefore no longer be to reconstruct information but to hallucinate small objects, which poses a problem in terms of learning quality. To avoid this problem, several papers focus on optimizing the masking strategy. \addSylv{The authors of}~\cite{liu2023pixmim} propose a conservative data transform to maintain clues about foreground objects. MST~\cite{li2021mst}, AttMask~\cite{kakogeorgiou2022hide} and AMT~\cite{liu2023good} rely on self-distillation and use attention maps derived from the teacher network to choose the regions to \addSylv{be} mask\addSylv{ed}. MST chooses to mask non-essential regions only, with a low masking ratio ($1/8$), while AttMask shows that masking important features at a moderate ratio ($10-50\%$) improves the fine-tuning performance. AMT also relies on attention-driven masking, however they use \addSylv{the} feature maps derived from MAE or SimMIM last layer attention head (thus they do not rely on siamese architecture for training) after a warm-up phase ($40$ epochs). 
\addSylv{Like} in~\cite{kakogeorgiou2022hide}, AMT makes the most informative parts more \addSylv{likely} to be masked although there is still a probability that they remain visible. The authors also show that not using middle attention patches increases the performance, while also reducing the training cost. MILAN~\cite{hou2022milan} also proposes \addSylv{a} semantic aware sampling by using attention maps derived from CLIP weights~\cite{radford2021learning} (joint text-image SSL pre-training). 
However, \addSylv{in contrast to} AttMask and AMT, a high probability of remaining unmasked is given to highly informative parts. This is motivated by two elements: i)~masking all representative parts of an image leads to very long pre-training, and ii)~\addSylv{due to} the specific design of their decoder (MAE-like decoder but with frozen representations of the unmasked patches, discussed later), the features extracted by the network need to be informative enough. Indeed, the network should not learn to ``hallucinate'' objects. 

\addSylv{The m}ethods presented so far allow for the decomposition of an image into informative and less informative parts (often foreground/background), and \addSylv{the} relations\addSylv{hips} between those parts (being intra or inter relationships) are learned by the network. What if we further decompose the image by introducing more semantic parts? SemMAE~\cite{li2022semmae} proposes semantic-aware adaptative masking strategy by using \addSylv{some} segmentation maps. These segmentation maps are learned in a SSL way by solving a reconstruction task where the targets are patches extracted by a pre-trained ViT (e.g., iBOT), and by adding a diversity constraint on the attention maps. The attention maps obtained are then used for segmenting the image into several semantic parts. The semantic-guided masking of SemMAE then consists \addSylv{of} progressively masking $75\%$ of each part (intra-part or local feature learning) at the beginning of the training, to masking $75\%$ of the parts (inter-part relationships) at the end of the training process. 

\addSylv{Based on} all the masking strategies presented so far, \addSylv{the} authors seem to agree on the following conclusions: i)~a high masking ratio is recommended to ensure meaningful representation learning, and ii)~a carefully designed masking strategy (using either attention or semantic maps) further improves the performance. However, it is not clear which parts should be masked. DPPMask~\cite{xu2023dppmask} may provide an answer that gets everyone on the same page: keep as much representative and diverse information in unmasked patches, while masking at a high ratio (e.g.\addSylv{,} $75\%$). Representative and diverse patches are selected using Determinantal Point Processes (DPP), which aim at reducing the semantic change of an image after masking (miss-alignment problem). It consists in computing the distance (using a Gaussian kernel, which depends on the Euclidean distance between the intensity values of \addSylv{the} patch pairs) between each patch and selecting those that are dissimilar from a selected subset. \addSylv{Due to the} computational complexity \addSylv{resulting} from the exact DPP formulation (matrix decomposition), DPPMask proposes a greedy approximation of DPP. DPPMask shows significant improvements over AttMask and SemMAE masking strategies for both MAE and iBOT.

\subsubsection{Reconstruction targets}

Although masking strategy is very important to improve the performance, there are also many discussions about the choice of the reconstruction targets.
First MIM methods~\cite{vincent2008extracting, he2022masked, xie2022simmim} attempt to reconstruct raw pixels, and apply the Mean Squared Error (MSE) loss as the reconstruction objective. 
\addSylv{An} important limitation with such a reconstruction objective is that all reconstructed pixels have the same weight in the loss, \addSylv{al}though some reconstruction errors may be irrelevant for meaningful feature extraction. 

\addSylv{Therefore, s}ome methods propose to adapt the reconstruction target to the downstream objectives. For example, to force the network to focus on shapes rather than texture and rich details, PixMIM~\cite{liu2023pixmim} filters the high frequencies in the target objective (and thus the network focuses on low frequencies). Ge$^2$-AE~\cite{liu2023devil} and A$^2$MIM~\cite{li2023architecture} apply the reconstruction loss in both spatial and frequency domains to learn global features. In the same spirit, MaskFeat~\cite{wei2022masked} uses the Histogram of Oriented Gradients (HOG) as a reconstruction target, and justify this choice by the fact that HOG provides local shapes and appearances while being invariant to photometric changes. 
In the same line, SSM~\cite{huang2022self} applies different reconstruction losses which introduce some global criteria that do not suppose independence between neighboring pixels, such as Gradient Magnitude Similarity (GMS) and Structured Similarity Index \addSylv{Measure} (SSIM). 

However, all these methods rely on computationally expensive architectures in order to reconstruct \addSylv{the} full-resolution (or almost) image, with a decoder that will not be used for \addSylv{the final task}.
To address this issue, some authors propose to reconstruct \addSylv{some} features instead of full-resolution images. In this case, the challenge consists in defining relevant target features. \addSylv{Among the ideas proposed and tested, the literature has retained}
\begin{itemize}
    \item \textbf{Features from a pre-trained network -- } Several methods\addSylv{, such as} BEiT~\cite{bao2021beit}, MaskDistill~\cite{peng2022unified}, MILAN~\cite{hou2022milan} and MaskAlign~\cite{xue2023stare}, rely on distillation 
    from strong unsupervised pre-trained encoders, such as CLIP. However\addSylv{,} such \addSylv{a} strategy may not be optimal on datasets that present a domain gap (e.g.\addSylv{,} satellite data) with, for example, CLIP pre-training data.
    \item \textbf{Features obtained via self-distillation -- } Another way to obtain target features is by relying on self-distillation methods and asymmetric siamese networks. Such \addSylv{a} strategy is adopted by SOTA methods like SplitMask~\cite{el2021large}, iBOT and I-JEPA~\cite{assran2023self}. I-JEPA differs from iBOT by the fact that \addSylv{it} ask\addSylv{s} the network to reconstruct not the full masked areas\addSylv{, but only parts} of the masked image given a context. Nonetheless, \addSylv{the} target features obtained using pre-trained weights seem to lead to better representation learning. \cite{liu2022exploring} claims that it is not necessary to carefully choose the target (HOG, MaskFeat or features obtained with MAE/SimMIM etc.) as long as \addSylv{a} multi-stage distillation pipeline is used, which leads to dBOT method. However, even with dBOT framework, CLIP pre-trained teacher still leads to better performance than \addSylv{a} randomly initialized teacher.
\end{itemize}

In the literature, many questions have been raised about the design of the decoder in the SSL pre-training phase. Some \addSylv{authors} argue that it is better to use a simple decoder to maximize transfer learning performance~\cite{xie2022simmim}, while others have observed that a deep and narrow decoder works best~\cite{he2022masked}. This is one of the open questions in the field of image reconstruction. Indeed, how can we ensure that it is the encoder \addSylv{and not the decoder} that learns to extract highly representative information from an image and to disentangle causal factors? 
MILAN~\cite{hou2022milan} proposes to circumvent this issue by designing a specific decoder that clearly separates the functional roles of the encoder and the decoder. \addSylv{To this end}, the authors introduce a prompting decoder that takes as input frozen representations of encoded unmasked patches. The latter are therefore used as fixed prompts. However, the ablation study  \addSylv{shows} that the SOTA performance achieved with MILAN is mainly due to the use of CLIP targets \addSylv{and} not to the design of the prompting decoder.

\subsubsection{Adaptation to convolutional networks}
\addSylv{M}ost papers tackling MIM rely on the use of ViT encoders. CNN-based encoders are still widely used in many real-world applications and have some advantages, such as faster inference times on small inputs, and a hierarchical architecture that benefits object detection. However, they seem to be less efficient than ViT encoders when combined \addSylv{with} MIM methods. This \addSylv{may be due to} their poor ability to estimate large-scale relationships between image patches. Also, unlike ViT architectures that analyze each patch independently, CNN-based encoders perform convolutions by sliding a window, and thus the receptive field of the convolution can overlap with both masked and unmasked areas. This leads to several issues such as masked pattern vanishing or the disturbance of \addSylv{the} distribution \addSylv{of} pixel values, as explained in~\cite{tian2022designing}.
A$^2$MIM~\cite{li2023architecture} attempts to solve this issue by replacing \addSylv{the} $0$-padding \addSylv{with a padding the} mean value of \addSylv{the} unmasked pixels. ConvMAE~\cite{gao2022convmae}, MixMAE~\cite{liu2023mixmae} and SparK~\cite{tian2022designing} introduce the use of partial or sparse convolution\addSylv{s}. Specifically, \addSylv{the} authors of \addSylv{the} SparK~\cite{tian2022designing} paper show that MAE pre-training with a CNN-based encoder can outperform ViT-based MAE pre-training when using sparse convolution and a modern CNN-based encoder, namely ConvX-B~\cite{liu2022convnet}. Furthermore,~\cite{liu2023mixmae} efficiently encodes two images as a single image by replacing \addSylv{the} masked patches of the first image (image 1) \addSylv{with} the unmasked ones of the second image (image 2). To adapt this strategy to ConvNets, they introduce unmixed convolutions, which consists in unmixing the image into image 1 and image 2, and then applying partial convolution.


\section{Benchmark on the COCO dataset}

Now that we have introduced the key SSL strategies for enhancing object detection, we propose to evaluate a selection of them on two benchmark datasets. \addSylv{The considered} SSL methods \addSylv{are} summarized in Table~\ref{tab:ssl_benchmark_strat}, \addSylv{and the} object detection framework \addSylv{corresponds to the} well-established and widely-used dataset from the literature, namely the COCO dataset~\cite{lin2014microsoft}. 
Various sizes of objects are covered, including $41\%$ of small objects (i.e., objects having an area lower than $32^2$ pixels). In the literature, although a large number of SSL papers evaluate their methods on the COCO dataset, the fine-tuning set-ups or \addSylv{the} evaluation conditions \addSylv{may} differ from one paper to another. To ensure a fair comparison,  we propose to fine-tune the studied SSL methods ourselves, using training parameters from recent papers that have prove\addSylv{n} their efficiency.

\bgroup
\def\arraystretch{1.15}
\begin{table}[ht] 
\small
\centering
 
  \begin{tabular}{cp{1.5cm}cc} 
    \textbf{Method} & \textbf{Category} & \textbf{Backbone} & \textbf{\#params} \\
    \hline
    
    \hline
    DINO~\cite{caron2021emerging} & Inst. Discr. (global) & R50 & 23M \\
    & & ViT/S-16 & 21M \\
    & & ViT/B-16 & 83M \\
     \hline

    ReSim~\cite{xiao2021region} & Inst. Discr. (local) & R50 & 23M \\
      \hline
      Leopart~\cite{ziegler2022self} & Inst. Discr. (local) & ViT/S-16 & 21M \\
      \hline
      SparK~\cite{tian2022designing} & MIM & R50 & 23M \\
    & & R200 & 65M \\
    \hline
    MAE~\cite{he2022masked} & MIM & ViT/S-16 & 21M \\
    & & ViT/B-16 & 83M \\
   
  \end{tabular}
  \caption{Compared pre-training methods, along with their SSL category, considered backbones and number of parameters in \addSylv{each} backbone. R50 stands for ResNet-50 backbone, R200 for ResNet-200, ViT/S-16 for Vision Transformer (ViT) Small version with a patch size of $16$, and ViT/B-16 for ViT Base version with a patch size of $16$. ``Inst. Discr.'' stands for instance discrimination methods and ``MIM'' for masked image modeling. }
  \label{tab:ssl_benchmark_strat}
\end{table}
\egroup


\subsection{Experimental set-up}
\label{exp_setup_coco}
We consider a Mask R-CNN~\cite{he_mask_2018} with ResNet-50 (R50), ResNet-200 (R200), ViT/B-16 or ViT/S-16 encoders as our detectors.  For the encoder, the pre-trained weights of each SSL method are taken from the Github repository published by the authors of the original papers. The fine-tuning parameters for the ResNet-based encoders (namely R50 and R200) are chosen following SparK's paper~\cite{tian2022designing} recommendations. More \addSylv{specifically}, we train the detector using AdamW optimizer~\cite{loshchilov2017decoupled} and the $3\times$ schedule (i.e., we trained the network for $3\times 12$ epochs). For the learning rate, \addSylv{since} we can only load $36$ images on our GPUs, we use the linear scaling rule introduced in~\cite{goyal2017accurate} to choose an appropriate learning rate. We consider the ``Step LR'' scheduler, and multiply the learning rate by $0.2$ \addSylv{at} epochs $3\times 9$ and $3\times 11$. For ViT-based fine-tuning, we follow the training set-up proposed in~\cite{fang2023unleashing} and scale the learning rate according to our GPU resources (four Nvidia A100 GPUs) based on the linear scaling rule. We fine-tune the neural networks for $50$ epochs using AdamW optimizer and CosineLR scheduler. 

We use ``COCO 2017 val'' subset as our test set and evaluate the box location accuracy of each method using the conventional mean average precision metric $\textmd{m}\textsc{AP}^{box}_{@0.5:0.95}$ (i.e., the area under the precision-recall curve, averaged over all the object classes and over $10$ IoU threshold from $0.5$ to $0.95$). In order to focus on the detection performance, we will also provide the metrics for box location regardless of the errors made on the classification ($\textsc{AP}^{box}_{@0.5:0.95}$). We will also focus on small object detection performance by providing \addSylv{these metrics} for objects that have a spatial extent \addSylv{less} than $32\times 32$ pixels ($\textsc{AP}^{box,S}_{@0.5:0.95}$, $\textsc{AP}^{box,S}_{@0.3}$). Since a small deviation in the box localization for small objects drastically reduces the IoU between the predicted box and the ground-truth, we introduce more tolerance regarding the localization errors by lowering the IoU threshold to $30\%$ ($\textsc{AP}^{box}_{@0.3}$, $\textsc{AP}^{box,S}_{@0.3}$). 

\subsection{Reproducibility}

First of all, we would like to make a few comments about the reproducibility of the results presented in the original papers. 

For the methods trained with a ResNet-50 encoder, the results \addSylv{we have obtained} are slightly better than those presented in the original papers. 
This difference \addSylv{can be} explained by the choice of a longer schedule, along with a different optimizer\addSylv{, namely} AdamW optimizer instead of the classical SGD optimizer.

For ViT-based fine-tuning, the results \addSylv{we have obtained} are worse than those reported in the original papers. For example, ~\cite{he2022masked} \addSylv{achieves} a $\textmd{m}\textsc{AP}^{box}_{@0.5:0.95}$ of $50.3\%$ on the COCO dataset using MAE pre-trained weight, while we can only achieve a $\textmd{m}\textsc{AP}^{box}_{@0.5:0.95}$ of $47.8\%$ ($-2.5\%$). 
This \addSylv{can be} partly explained by the fact that we considered a shorter fine-tuning schedule (only $50$ epochs instead of $100$ epochs in~\cite{he2022masked}). Moreover, since we did not have access to the same amount of GPU resources as the original papers, we were forced to drastically reduce the size of our batches. Despite adapting the learning rate accordingly, it is likely that the linear scaling rule~\cite{goyal2017accurate} does not \addSylv{directly} apply, meaning that our training parameters are not optimal. Due to the excessive computation time, the search for optimal training parameters has been set aside, and it must therefore be assumed that there is a slight difference in the results, of about $2\%$ or $3\%$.

\subsection{Results}

\begin{table*}[t] 
\small
  \centering
\begin{tabular}{ccccccc} 
  \hline
    \multicolumn{2}{c}{\multirow{2}{*}{Backbone}}  & With Class. & \multicolumn{2}{c}{No Class.} & \multicolumn{2}{c}{Small objects, no Class.}\\
     & & $\textmd{m}\textsc{AP}^{box}_{@0.5:0.95}$ & $\textsc{AP}^{box}_{@0.5:0.95}$  & $\textsc{AP}^{box}_{@0.3}$ & $\textsc{AP}^{box,S}_{@0.5:0.95}$  & $\textsc{AP}^{box,S}_{@0.3}$\\ 
     \hline
     \multicolumn{6}{l}{\textbf{Small networks (21-23 M \#params.)}}\\
     \hline
    \multicolumn{4}{l}{\textcolor{darkgray}{\textit{Instance discrimination methods}}}  \\
     DINO & R50 & 42.8 & 46.9 &  77.0 &31.9 &  64.0 \\
     ReSim & R50 & 44.3 & 48.6 &78.4 & \textbf{33.3} &  65.7\\
     DINO & ViT/S-16 & \underline{46.3} & \underline{48.8} & \underline{79.9} & 32.2  & \underline{66.9}\\
     Leopart & ViT/S-16 & \textbf{46.5} &\textbf{49.0} & \textbf{80.1} & 32.4 &  \textbf{67.1}\\
      \hline
    \textcolor{darkgray}{\textit{MIM methods}}  & & & & &\\
    SparK & R50 & 44.1 & 48.6 & 78.0 &\textbf{33.3} &  64.9 \\
\hline
     \multicolumn{6}{l}{\textbf{Large networks ($\geq$ 65 M \#params.)}}\\
     \hline
    \multicolumn{4}{l}{\textcolor{darkgray}{\textit{Instance discrimination methods}}}  \\
     DINO &  ViT/B-16 & \underline{47.3} & 49.1 & \underline{80.2} &32.7&  \textbf{68.1}\\
     \hline
     \textcolor{darkgray}{\textit{MIM methods}}  & & & & & \\
     SparK & R200 & 46.7 & \textbf{50.5} & 79.4 &\textbf{35.2} &  67.5\\
     MAE & ViT/B-16 & \textbf{47.8} & \underline{50.3} &  \textbf{80.7} &\underline{33.4}& \underline{67.6} \\
     
   \hline
  \end{tabular}
  \caption{Benchmark on the COCO dataset with (``With Class.'') or without classification labels (``No Class.'', i.e., detection \addSylv{only}). For each network size (small or large), \addSylv{the} best results are in bold and \addSylv{the} second best results are underlined. }
  \label{tab:coco_benchmark_detection}
\end{table*}

Table~\ref{tab:coco_benchmark_detection} \addSylv{presents} the results obtained on COCOval 2017 dataset. Our observations are the following:

\paragraph{The encoder architecture matters more than the SSL strategy} 
\addSylv{According to} Table~\ref{tab:coco_benchmark_detection}, large networks, especially those based on ViT/B-16 backbone, lead to the best results. For example, the $\textmd{m}\textsc{AP}^{box}_{@0.5:0.95}$ is increased by $2.6\%$ when considering a ResNet-200 encoder instead of a ResNet-50 encoder for SparK, and increased by $1\%$ when considering a ViT/B encoder instead of a ViT/S for DINO. Note that the performance gap is narrower for ViT encoders \addSylv{than with CNN}. Moreover, ViT backbones perform significantly better than ResNet backbones.
However, the performance gap is reduced \addSylv{if} classification errors \addSylv{are ignored}, especially when it comes to small objects. Indeed, Table~\ref{tab:coco_benchmark_detection} shows that SparK initialization on a ResNet-200 encoder leads to an $\textsc{AP}^{box,S}_{@0.5:0.95}$ that is $1.8\%$ better than MAE initialization on a ViT/B-16. \addSylv{We deduce} that ResNet encoders are likely to be more prone to classification errors than ViT encoders.
    
\paragraph{Introducing locality in the SSL pre-training is important for ResNet-based encoders} Let us now take a closer look at the performance obtained by each SSL strategy. Concerning ResNet-50 backbone, it is clear that ReSim outperforms the other pre-training strategies.
SparK (MIM method) leads to competitive performance, while DINO seems to be the worst SSL training strategy for this task. The results seem \addSylv{to be} consistent with our intuition: \addSylv{in contrast to} global instance discrimination, \addSylv{both} local instance discrimination and MIM methods force the \addSylv{neural} networks to model local interactions within the image, which may benefit object detection. When looking at the detection performance only (i.e., no classification), we notice that, for ResNet-50 backbones, ReSim and SparK lead to very close results even on small objects, although ReSim is slightly better than SparK when lowering the IoU threshold. The performance gap with DINO remains very large, especially for small objects. 

\paragraph{ViT encoders are less sensitive to the pre-training strategy} For ViT encoders, the difference in performance between the SSL strategies is very thin: although local methods (MIM or local instance discrimination methods) seem to perform slightly better in terms of $\textsc{AP}^{box}_{@0.5:0.95}$, introducing more tolerance towards localization errors shows that DINO with ViT/S-16 or ViT/B-16 encoder is also very competitive on small object detection. Furthermore, DINO with ViT/B-16 encoder leads to the best $\textsc{AP}^{box,S}_{@0.3}$ score. This suggests that, in an ideal and data-sufficient case, \addSylv{the} ViT backbones are less sensitive to the pre-training strategy compared to \addSylv{the} ResNet encoders.

\paragraph{ViT encoders are more prone to localization errors on small objects} \addSylv{Still referring to Table~\ref{tab:coco_benchmark_detection}}, the $\textsc{AP}^{box,S}_{@0.5:0.95}$ column shows that ResNet-based encoders lead to \addSylv{the} best performance on small objects (e.g., $+0.9\%$ in $\textsc{AP}^{box,S}_{@0.5:0.95}$ when comparing Leopart and ReSim), meaning that these architectures are better suited \addSylv{to} small object detection. Nevertheless, the introduction of greater tolerance to localization errors reveals that ViT encoders are still capable of detecting small objects, \addSylv{albeit} with a slightly worse localization accuracy.

\section{What about domain-specific tasks?}

In this section, we challenge the previously studied SSL pre-training strategies in a real-\addSylv{world} scenario, namely small vehicle detection from remote sensing data. For this purpose, we consider the VEDAI dataset~\cite{razakarivony2016vehicle}, which is composed of $1200$ RGB and IR satellite scenes \addSylv{containing} small vehicles. This allows us to study the cross-domain transfer ability of the considered pre-training \addSylv{strategies, from RGB} to IR domain. We will try to answer the following questions: 1)~does SSL pre-training benefit real-world small object detection? 2)~is it better to perform SSL pre-training on a dataset whose statistics are close to those of the target data? (e.g., infrared dataset, remote sensing data), 3)~which SSL strategy is best for pre-training on an uncleaned dataset (i.e., with high temporal redundancy, low diversity, etc.)?, and 4) can SSL benefit few-shot training?

\subsection{Experimental set-up}
\begin{table}[t] 
\small
  \centering
\begin{tabular}{cccccc} 
  \hline
  \multicolumn{2}{c}{\multirow{2}{*}{Backbone}} & \multicolumn{2}{c}{VEDAI RGB} & \multicolumn{2}{c}{VEDAI IR}\\
     & & $\textsc{AP}$ & F1 & $\textsc{AP}$ & F1 \\ 
     \hline
     \multicolumn{4}{l}{\textbf{Small networks (21-23 M \#params.)}}\\
     \hline
     Scratch & R50 & 61.8 & 62.5 & 61.3 & 60.9 \\
      Scratch & ViT/S-16 & 79.4 & 72.8 & 74.8 & 71.3 \\
    
     \hline
    \multicolumn{4}{l}{\textcolor{darkgray}{\textit{Instance discrimination methods}}}  \\
     DINO & R50 & 86.1 & 82.0 & 84.0 & 79.0 \\
     ReSim & R50 & 87.7 & 84.4 & \underline{85.1} & \underline{81.6} \\
     DINO & ViT/S-16 & 89.7 & 81.8 & 84.4 & 78.1\\
     Leopart & ViT/S-16 & \underline{91.0} & \underline{84.5} & 84.3 & 78.0 \\
      \hline
     \multicolumn{4}{l}{\textcolor{darkgray}{\textit{MIM methods}}} \\
    SparK & R50 & 86.4 & 83.2 & 81.1 & 78.4\\
    MAE & ViT/S-16 & \textbf{91.8} & \textbf{86.1} & \textbf{88.4} & \textbf{83.7} \\
\hline
     \multicolumn{4}{l}{\textbf{Large networks ($\geq$ 65 M \#params.)}}\\
     \hline
     Scratch & ViT/B-16 & 66.7 & 63.2 & 58.5 & 57.3 \\
      \hline
     \multicolumn{4}{l}{\textcolor{darkgray}{\textit{Instance discrimination methods}}}  \\
     DINO &  ViT/B-16 & \textbf{94.9} & \textbf{89.6}  & \textbf{90.7} & \underline{85.6}\\
     \hline
    \multicolumn{4}{l}{\textcolor{darkgray}{\textit{MIM methods}}} \\
     MAE & ViT/B-16 & \underline{94.1} & \underline{88.5}  & \textbf{92.1} & \textbf{86.0}\\
     
   \hline
  \end{tabular}
  \caption{Benchmark of different pre-training methods on the VEDAI RGB and IR datasets. For each network size (small or large), \addSylv{the} best results are in bold and \addSylv{the} second best results are underlined.}
  \label{tab:vedai_rgb}
\end{table}

We fine-tune a Faster R-CNN on the RGB version of the VEDAI dataset with various encoders initialized with different pre-training strategies (SSL or supervised on ImageNet). The training parameters are \addSylv{those} used in Section~\ref{exp_setup_coco}, except that we considered a CosineLR scheduler for ResNet-based architectures since it leads to better performance. We split the VEDAI dataset into training, validation and test sets using a ratio of $60:20:20$, and consider the AP (with an IoU threshold of $5\%$) and F1 score metrics for evaluation. Since ViT/S-16 weights pre-trained using MAE strategy are not available in the literature, we decided to perform MAE pre-training on ImageNet dataset ourselves. We used the same training parameters as in the original paper and trained the encoder for $400$ epochs. 

\subsection{Results obtained on the VEDAI RGB dataset}
\addSylv{According to Table~\ref{tab:vedai_rgb}}, on the RGB version of VEDAI dataset, there is a large gap between the performance obtained using a ResNet-50 and a ViT encoder. In particular, \addSylv{the use of} large ViT encoders lead\addSylv{s} to impressive performance on this dataset. For example, a ViT/S-16 encoder can achieve an AP of almost $92\%$, while ResNet encoders merely reach an AP of $87.7\%$. 
Let us now dive into the performance achieved by the different SSL strategies. For ResNet-50 backbones, ReSim pre-training performs significantly better than DINO and SparK pre-training \addSylv{strategies}. For ViT backbones, it is \addSylv{difficult} to draw conclusion\addSylv{s}: MAE seems to benefit the most \addSylv{for} small encoder pre-training, while DINO \addSylv{performs} slightly better than MAE \addSylv{with} a larger encoder. It seems that, for ViT encoders, the fine-tuning performance on the final task is less dependent on the ViT initialization, which is in line with what was observed on the COCO dataset. 
In the end, it seems that the choice of a good encoder, especially those based on ViT blocks, is more important for the performance of the downstream task than the choice of a good pre-training strategy. But what if we consider a downstream task dataset whose image statistics are very different from those of ImageNet? 

\subsection{Transferring the knowledge learned on RGB data to IR domain}
We now evaluate the ability of the different pre-training strategies to transfer to other spectral domains \addSylv{using} IR imagery \addSylv{as a target example}. For this purpose, we consider the IR images of VEDAI dataset and coined this subset of data as VEDAI IR. We fine-tune a Faster R-CNN with different pre-trained encoders in the same way as previously. Note that these encoders have been pre-trained on RGB images (ImageNet dataset). The last two columns of Table~\ref{tab:vedai_rgb} \addSylv{show} the results obtained on VEDAI IR dataset.  We first notice that there is a large drop in performance for ViT-based instance discrimination pre-training \addSylv{strategies}, and they perform even wors\addSylv{e} than the ResNet-based pre-trainings (for equivalent network size). Indeed, DINO and Leopart pre-training \addSylv{strategies} with ViT/S-16 perform about $5\%$ worse in $\textsc{AP}^{box}_{@0.05}$ when applied to VEDAI IR dataset, while MAE leads to a decrease of only $2\%$. The performance gap is less pronounced when it comes to larger networks, and MAE leads to the best performance. 

For ResNet backbones, ReSim seems to be significantly more robust than any other pre-training \addSylv{strategy}, while SparK suffers from a large drop in performance. 
 \addSylv{According to} these observations\addSylv{,} the fine-tuning performance of SSL pre-trained weights varies greatly depending on the encoder architecture considered: MIM methods combined with ViT encoders seem to generalize better to datasets that statistically differ from the ImageNet dataset, whereas in the case of ResNets, it is the instance discrimination methods that perform best.
This may be explained by the fact that MIM methods are very sensitive to the image statistics, due to their strong bias towards local details (e.g.\addSylv{,} textures), and may therefore show a decrease in performance when applied to a different dataset. However, since ViT encoders are better at modeling large-scale dependencies (i.e.\addSylv{,} they have a bias towards shapes), the combination of ViT encoders and MIM methods compensates for the weakness observed for the latter. Thus the following question arises: can we improve the performance by pre-training on a dataset that has close characteristics to the downstream task dataset? 
To answer this question, we perform some SSL pre-training on an  infrared dataset\addSylv{, that however is uncleaned (i.e., without removal of redundant images)}. \addSylv{Results are commented in the next paragraph.} This will also allow us to assess the degree of generalization ability of SSL pre-training to other pre-training databases.


\subsection{Pre-training on an uncleaned infrared dataset}
\label{sec:VEDAI_IR_SSL}
\addSylv{To be able to perform SSL pre-training on an infrared dataset}, we collected a \addSylv{large number} of infrared images from several publicly available infrared datasets. Table~\ref{ir_ssl_dataset_source} summarizes the different infrared dataset sources \addSylv{that} we merged together in order to obtain a large infrared dataset, and we coined the final dataset as \textbf{SSL-IR} dataset. The datasets we used to obtain SSL-IR have very different characteristics: they contain different scenes (urban, sky, forest...) \addSylv{captured} from various camera viewpoints (drone, car), and with different infrared sensors (thermal infrared, near infrared, etc.). \addSylv{However, m}ost images are extracted from video sequences, and thus the obtained dataset suffers from low \addSylv{image} diversity. We obtain a total of approximately $720$k infrared images, which represents about $60\%$ of ImageNet-1k dataset. 
\bgroup
\def\arraystretch{1.15}
\begin{table*}[ht] 
\small
\centering
 
  \begin{tabular}{cp{6cm}p{2.5cm}c} 
    \textbf{Source dataset} & \textbf{Type of data} & \textbf{Nature of data} & \textbf{\# images} \\
    \hline

    \hline
    
      LSOTB-TIR~\cite{liu2020lsotb} & drone, car, fixed cameras, urban sky natural scenes, thermal infrared object tracking & video & 524k \\
      \hline
      IRDST~\cite{sun2023receptive} & real and simulated data, drone, sky and urban scene, target detection & video & 143k \\
      \hline
      FLIR~\cite{flir-data-set_dataset} & car, urban scenes, autonomous driving & video & 35k \\
      \hline
    MFIRST~\cite{wang2019miss} & drone, sky and urban scenes, simulated and real small target detection  & single-frame images & 10k \\
    \hline
    ASL-TID~\cite{portmann2014people} & drone, urban scenes, pedestrian detection & video & 4k \\
    \hline
    HIT-UAV~\cite{suo2023hit} & drone, urban scenes, pedestrian detection & video & 3k \\
    \hline
    IRSTD-1k~\cite{zhang2022isnet} & drone, sky, natural and urban scenes, small target detection & single-frame images & 1k \\
      \hline
      
     \hline
   \textbf{SSL-IR} & & & \textbf{720k}
  \end{tabular}
  \caption{SSL-IR dataset: data sources and specifications.}
  \label{ir_ssl_dataset_source}
  
\end{table*}
\egroup
\def\arraystretch{1.15}

\begin{table}[t] 
\small
  \centering
\begin{tabular}{llll} 
  \hline
     & Backbone & $\textsc{AP}^{box}_{@0.05}$ & F1 \\ 
     \hline
     
     \hline
     Scratch & R50 & 61.3 & 60.9 \\
    
     \hline
   \multicolumn{4}{l}{\textcolor{darkgray}{\textit{Instance discrimination methods}}}  \\
     ReSim-IR & R50 & $76.6^{\color{red}{(-8.5)}}$ & $72.9^{\color{red}{(-8.7)}}$ \\
     Leopart-IR & ViT/S-16 & $81.6^{\color{red}{(-2.7)}}$ & $76.7^{\color{red}{(-1.3)}}$ \\
      \hline
      \multicolumn{4}{l}{\textcolor{darkgray}{\textit{MIM methods}}} \\
    SparK-IR & R50 & $77.4^{\color{red}{(-3.7)}}$ & $75.0^{\color{red}{(-3.4)}}$ \\
    MAE-IR & ViT/S-16 & $\textbf{88.5}^{\color{red}{(-0.1)}}$ & $\textbf{82.8}^{\color{red}{(-0.9)}}$\\
   \hline
  \end{tabular}
  \caption{Benchmark on VEDAI IR with SSL methods pre-trained on SSL-IR dataset. \addSylv{The b}est results are in bold, and the performance gaps with the respective SSL strategies pre-trained on ImageNet are indicated in the superscript.  }
  \label{tab:vedai_ir_ssl_ir}
\end{table}

We pre-train\addSylv{ed} ReSim (R50), SparK (R50), Leopart (ViT/S-16) and MAE (ViT/S-16) on \addSylv{the} SSL-IR dataset using the pre-training parameters suggested for each method in the original papers. We then fine-tuned a Faster R-CNN on VEDAI IR  \addSylv{under} the same conditions as \addSylv{before}. The results are \addSylv{shown in} Table~\ref{tab:vedai_ir_ssl_ir}. \addSylv{According to this table,} ReSim suffers from a huge drop in performance (more than $8\%$ in both AP and F1 score), while the decrease in performance is limited for SparK and Leopart. Moreover, MAE is particularly robust to training on SSL-IR dataset, since the performance is almost equivalent to the pre-training on ImageNet. Overall, for both ResNet and ViT encoders, MIM-based SSL pre-training is more robust to pre-training on a smaller and less clean dataset than its instance discrimination counterparts. At first sight, the results of the pre-training on SSL-IR are rather disappointing compared to the RGB weights available in the literature. However, it should be remembered that the IR dataset we considered is not cleaned and is even much smaller than ImageNet. Furthermore, by choosing the right SSL strategy and encoder, we can obtain results that are very similar to those given by the weights in the literature. This is encouraging, especially in cases where it is absolutely necessary to pre-train SSL on custom datasets (for example, if there is a large domain gap, or if the encoder architecture needs to be significantly changed).

\subsection{Frugal setting}
\begin{table}[t] 
\small
  \centering
\begin{tabular}{cccccc} 
  \hline
  \multicolumn{2}{c}{\multirow{2}{*}{Backbone}} & \multicolumn{2}{c}{25-shots} & \multicolumn{2}{c}{50-shots}\\
     & & $\textsc{AP}$ & F1 & $\textsc{AP}$ & F1 \\ 
     \hline
     
     \hline
     Scratch & R50 & 30.1 & 22.2&33.9&25.9 \\
      Scratch & ViT/S-16 & 14.8&6.8&24.2&13.9 \\
    
     \hline
    \multicolumn{4}{l}{\textcolor{darkgray}{\textit{Instance discrimination methods}}}  \\
     DINO & R50 & \underline{38.2}&\underline{33.4}&\underline{53.4}&\underline{53.3} \\
     ReSim & R50 & \textbf{50.4}&\textbf{52.0}&\textbf{57.5}&\textbf{58.2} \\
     DINO & ViT/S-16 & 20.1&8.9&30.9&21.5\\
     Leopart & ViT/S-16 &17.7&9.5&30.3&21.9\\
      \hline
     \multicolumn{4}{l}{\textcolor{darkgray}{\textit{MIM methods}}} \\
    SparK & R50 & 34.4&29.2&48.8&44.7\\
    MAE & ViT/S-16 & 33.9&29.2&42.4&37.2 \\
\hline
     
  \end{tabular}
  \caption{Results obtained on VEDAI RGB in 25 and 50-shot settings. \addSylv{The} best results are in bold and \addSylv{the} second best results are underlined.}
  \label{tab:frugal_vedai}
\end{table}

Finally, we evaluate the different SSL strategies in challenging fine-tuning conditions, namely few-shot setting. For this purpose, we consider fine-tuning on $25$ or $50$ images from the VEDAI RGB dataset. The results are presented in Table~\ref{tab:frugal_vedai}. 
In general, we can see that there is a real contribution of using SSL pre-trained weights in few-shot setting, although the benefits are more or less obvious depending on the SSL strategies or architectures used.
Firstly, we can see that ViT/S encoders \addSylv{achieve} significantly inferior performance compared to ResNet-50, even when relying on SSL pre-trained weights. 
Secondly, the choice of the SSL pre-training strategy depends on the encoder. For ResNet-50, instance discrimination methods, in particular ReSim, significantly benefits $25$ and $35$-shot trainings. This is evidenced by an improvement of over $20\%$ in terms of AP and F1 score when compared to a network that has been trained from scratch. SparK exhibits only marginal improvement over randomly initialized weights, especially in the $25$-shot setting. Regarding the results obtained with a ViT/S encoder, a notable improvement is observed when ViT is combined with the MIM method (specifically MAE), although the performance remains inferior to that observed with ResNet-50.

\section{Conclusion}

In this paper, we presented a survey of SSL strategies oriented towards local feature extraction, which appear better suited \addSylv{to} object detection tasks. We performed a benchmark using two distinct datasets: 1)~the COCO dataset, which represents an ideal scenario for object detection with a large amount of diverse data, and 2)~the VEDAI dataset, a real-world, domain-specific case with IR images that deals with much smaller objects and more complex, diverse backgrounds, making it quite different from the ImageNet dataset used \addSylv{by the authors} to pre-train the\addSylv{ir} SSL methods. These benchmarks allowed us to draw important conclusions to guide future \addSylv{users} in choosing appropriate pre-training strategies based on their specific use cases. The key takeaways are:

\begin{itemize}
    \item \textbf{Importance of the encoder choice:} The selection of the encoder is more critical than the choice of the pre-training strategy. ViT encoders generally outperform ResNets when sufficient fine-tuning data is available and when dealing with large objects. In this case, ViTs are less sensitive to the pre-training strategy. However, they tend to perform poorly in frugal settings, and should be combined with MIM methods in such cases.

    \item \textbf{ResNets are more sensitive to the SSL pre-training:} ResNets perform better when combined with local instance discrimination methods or MIM. However, MIM pre-training leads to poor performance in a frugal setting.
    
    \item \textbf{Domain shift:} We observed that pre-training on in-domain images does not necessarily improve performance and may even degrade it in \addSylv{the considered case, namely from RGB to IR}. This might be because IR images are still relatively close to RGB, which explains why weights pre-trained on RGB data can generalize well to IR data.  However, conclusions might differ with more significant domain shifts (e.g., astronomy or medical images), and SSL pre-training on a custom (and maybe uncleaned dataset) may be necessary. In this case, MIM methods should be prioritized for both ViT and ResNet networks. Note however that if the downstream tasks deals with frugal dataset, combining MIM and instance discrimination as in  CMAE~\cite{huang2023contrastive} or Siamese image modelling~\cite{tao2023siamese} could yield better results for ResNets.
\end{itemize}


Future work should focus on providing more theoretical explanations for the differences in behavior of pre-trained SSL strategies depending on the encoder. For example, we hypothesize that because ViTs model long-range dependencies, they are well-complemented by local SSL methods \addSylv{such as} MIM. This hypothesis needs further investigation. Additionally, exploring other application areas, such as anomaly detection, could help expand and complete this benchmark.

\textbf{Acknowledgments --} This project was provided with computing HPC and storage resources by GENCI at IDRIS thanks to the grant 2023-AD011014896 on the supercomputer Jean Zay's V100 and A100 partitions. It was also performed using computational resources from the “Mésocentre” computing center of Université Paris-Saclay, CentraleSupélec and École Normale Supérieure Paris-Saclay supported by CNRS and Région Île-de-France.
\ifCLASSOPTIONcaptionsoff
  \newpage
\fi



\bibliographystyle{IEEEtran}
\bibliography{bibtex/bib/IEEEabrv}
%
%

\end{document}